\documentclass[11pt,a4paper]{article}
\usepackage[hyperref]{eacl2021}
\usepackage{times}
\usepackage{latexsym}

\usepackage{booktabs}
\usepackage{amsmath}
\usepackage[ruled,vlined]{algorithm2e}

\usepackage{xcolor}

\def\E{{\rm E}}
\def\d{\frac{\partial}{\partial \theta}}
\def\N{\mathcal{N}}

\usepackage{graphicx}
\usepackage{caption}
\usepackage{subcaption}
\usepackage{microtype}

\aclfinalcopy 

\title{Generative Text Modeling through Short Run Inference}

\newcommand*\samethanks[1][\value{footnote}]{\footnotemark[#1]}
\author{Bo Pang\thanks{\hspace{5pt}Equal contributions.} \\
  \normalsize{UCLA} \\
  \tt\small{bopang@ucla.edu} \\\And
  Erik Nijkamp\samethanks{} \\
  \normalsize{UCLA} \\
  \tt\small{enijkamp@ucla.edu} \\\And
  Tian Han \\
  \normalsize{Stevens Institute of Technology} \\
  \tt\small{than6@stevens.edu} \\\And
  Ying Nian Wu \\
  \normalsize{UCLA} \\
  \tt\small{ywu@stat.ucla.edu}}

\date{}

\begin{document}
\maketitle
\begin{abstract}
Latent variable models for text, when trained successfully, accurately model the data distribution and capture global semantic and syntactic features of sentences. The prominent approach to train such models is variational autoencoders (VAE). It is nevertheless challenging to train and often results in a trivial local optimum where the latent variable is ignored and its posterior collapses into the prior, an issue known as \textit{posterior collapse}. Various techniques have been proposed to mitigate this issue. Most of them focus on improving the inference model to yield latent codes of higher quality. The present work proposes a short run dynamics for inference. It is initialized from the prior distribution of the latent variable and then runs a small number (e.g., 20) of Langevin dynamics steps guided by its posterior distribution. The major advantage of our method is that it does not require a separate inference model or assume simple geometry of the posterior distribution, thus rendering an automatic, natural and flexible inference engine. We show that the models trained with short run dynamics more accurately model the data, compared to strong language model and VAE baselines, and exhibit no sign of posterior collapse. Analyses of the latent space show that interpolation in the latent space is able to generate coherent sentences with smooth transition and demonstrate improved classification over strong baselines with latent features from unsupervised pretraining. These results together expose a well-structured latent space of our generative model.
\end{abstract}

\section{Introduction}
The state-of-the-art language models (LM) are often modeled with recurrent neural networks (RNN) \citep{mikolov2010recurrent} or attention-based models \citep{unilm, vaswani2017attention}. They are optimized by making a series of next-step predictions, encouraging the models to capture local dependency rather than global semantic features or high-level syntactic properties. A seminal work by \citet{bowman-etal-2016-generating} extends the standard LM to incorporate a continuous latent space which is aimed to explicitly capture global features. They formulate and train the model as a varational autoencoder (VAE) \citep{kingma2013auto}. Indeed, the model is able to generate coherent and diverse sentences through continuous sampling, and provide smooth interpolation between sentences, uncovering a well-formed latent space.  

However, training VAE for text is challenging and often leads to a trivial local optimum, \textit{posterior collapse}. Specifically, the training objective of VAE can be decomposed into a reconstruction term and a KL term that regularizes the distance between the posterior and prior of the latent variable. Due to the autoregressive nature of the decoder, it is able to reconstruct the data well by simply relying on the one-step-ahead groud-truth and evolving model state while completely ignoring the latent codes. The posterior hence collapses into the prior, carrying no information. This is an important open question in this field. As pointed out in \citet{fu2019cyclical}, two paths work together to generate sentences in VAE. One path (Path A) is through the latent codes, while the other (Path B) is conditioned on the prior ground-truth or previously generated tokens. The posterior collapse describes an easy solution, that is, relying on Path B and ignoring Path A. Prior efforts made to address this issue by and large are along the two paths. One can control the information available from Path B to force the decoder to employ more Path A information. \citet{bowman-etal-2016-generating} dropout the input words to the decoder and \citet{yang2017dilated} utilize a dilated CNN to control the size of context from previously generated words. Along Path A, various techniques have been developed to improve the latent code quality. \citet{bowman-etal-2016-generating} anneal the weight of the KL term from a small number to reduce the regularization in the beginning of the training (Anneal-VAE), while \citet{fu2019cyclical} further propose to adopt a cyclical annealing schedule (Cyclical-VAE). \citet{he2018lagging} update the encoder multiple times before one decoder update (Lagging-VAE). \citet{li2019surprisingly} initialize the VAE with an autoencoder (AE) and adopt a hinge loss for the KL term such that KL is not driven down below a target rate (FBP-VAE and FB-VAE). These techniques fall under the framework of amortized variational inference. Despite its fast inference, \citet{cremer2018inference} observes that an amortization gap, the gap between the log-likelihood and the ELBO, can be large. Thus \citet{kim2018semi} proposes semi-amortized variational autoencoders (SA-VAE) in which initial variational parameters are obtained from an encoder as in VAE, and the ELBO is then optimized with respect to the variational parameters to refine them.

An alternative to variational inference is Markov chain Monte Carlo (MCMC) sampling. MCMC posterior sampling may be in the form of Langevin dynamics~\cite{langevin1908theory} or Hamiltonian Monte Carlo (HMC)~\cite{neal2011mcmc,chen104hmc}. Traditional MCMC can be time-consuming as the Markov chains require a long running time, each iteration involving a gradient computation through the decoder.

In this article, we propose to apply a short run inference (SRI) dynamics, such as finite step Langevin dynamics, guided by the posterior distribution of the latent variable as an approximate inference engine. For each training example, we initialize such a short run dynamics from the prior distribution such as Gaussian noise distribution, and run a finite number (e.g., 20) of steps of updates. This amounts to a residual network which transforms the initial noise distribution to an approximate posterior distribution. 

One major advantage of the SRI is that it is natural and automatic. Designing and tuning a separate inference model is not a trivial task. In prior work, the inference model requires careful tuning to avoid posterior collapse in VAEs for text modeling. For instance, the inference model needs to be aggressively trained \citep{he2018lagging}, pre-trained with an autoencoder \cite{li2019surprisingly}, or refined with gradient descent guided by the ELBO \citep{kim2018semi}. In contrast, the short run dynamics guided by the log-posterior of the latent variable can be automatically obtained on modern deep learning platforms. In addition, our method does not assume a closed-form density for the posterior, like a Gaussian with diagonal covariance matrix, and hence are possible to have a good approximate posterior and provide good latent code. Lastly, we optimize the hyper-parameter of the short run dynamics by minimizing the KL divergence between the short-run-dynamics-induced posterior and the true posterior, to further improve the approximate posterior.

Empirically, we show that the model trained with the SRI is able to outperform a standard LSTM language model by employing an LSTM generative model, while exhibiting active utilization of the latent space, improving over models trained with VAE-based approaches. Moreover, we find the learned latent space is smooth, allowing for coherent and smooth interpolation and reconstruction from noisy samples, and captures sufficient global information, enabling enhanced classification accuracy over state-of-the-art baselines. 

In summary, the following are contributions of our paper. (1) We propose to use short run inference dynamics to train generative models for sentences without the need for an auxiliary inference network. (2) We demonstrate that the generative model trained with the SRI is able to accurately model the data distribution and make active use of the latent space, exhibiting no sign of posterior collapse. (3) We show that the learned latent space is smooth and captures rich global representations of sentences.

\section{Model and learning algorithm}

\subsection{Generative model}
Let $x$ be the observed example, such as a sentence. Let $z$ be the latent variable. We may consider $z$ as forming an interpretation or explanation of $x$, such as the global semantics and/or high-level syntactic properties of sentences. Consider the following generative model for $x$,
\begin{align}
    z \sim p(z) \quad x \sim p_{\theta}(x | z).
\end{align}
where $p(z)$ is the prior and $p_{\theta}(x | z)$ is given by a generative model parameterized with $\theta$. The marginal distribution of $x$ is 
$
p_\theta(x) = \int p_\theta(x, z) dz. 
$
Given $x$, the inference of $z$ can be based on the posterior distribution $p_\theta(z|x) = p_\theta(x, z)/p_\theta(x)$. 

\subsection{Learning and inference} 

Let $p_{\rm data}(x)$ be the data distribution that generates the example $x$. The learning of parameters $\theta$ of $p_\theta(x)$ can be based on 
$
\min_\theta {\rm KL}(p_{\rm data}(x)\| p_\theta(x)),
$
where ${\rm KL}(p\|q) = \E_p[\log (p(x)/q(x))]$ is the Kullback-Leibler divergence between $p$ and $q$ (or from $p$ to $q$ since ${\rm KL}(p\|q)$ is asymmetric). If we observe training examples $\{x_i, i = 1, ..., n\} \sim p_{\rm data}(x)$, the above minimization can be approximated by maximizing the log-likelihood 
\begin{align}
L(\theta)  = \frac{1}{n} \sum_{i=1}^{n} \log p_\theta(x_i), 
\end{align}
which leads to the maximum likelihood estimate (MLE). 

The gradient of the log-likelihood, $L'(\theta)$, can be computed according to the following identity: 
\begin{align} 
    \d \log p_\theta(x) &= \frac{1}{p_\theta(x)} \d p_\theta(x) \nonumber \\ 
    &= \frac{1}{p_\theta(x)} \int \d p_\theta(x, z) dz  \nonumber \\ 
    &= \int \d \log p_\theta(x, z) \frac{p_\theta(x, z)}{p_\theta(x)} dz \nonumber \\ 
    &= \E_{p_\theta(z|x)} \left[ \d \log p_\theta(x, z)\right]. 
\end{align}
While the marginal distribution ${p(x) =\int p(x|z)p(z)dz}$ is intractable due to the latent variables $z$ being integrated out, the above expectation can be approximated by Monte Carlo average with samples drawn from $p_\theta(z|x)$. Such samples from $p_\theta(z|x)$ can be obtained by MCMC in the form of Langevin dynamics~\cite{langevin1908theory}, which iterates 
\begin{align} 
   z_{k+1} = z_k + s \frac{\partial}{\partial z} \log p_\theta(z_k|x) + \sqrt{2s} \epsilon_k,
\label{eq:langevin}
\end{align}
where $\epsilon_k \sim \N(0, I)$, $t$ denotes the time step of Langevin dynamics, and $s$ is the discretization step size. The gradient term is tractable since $\frac{\partial}{\partial z} \log p_\theta(z_k|x) = \frac{\partial}{\partial z} \log p_\theta(z_k, x)$ and thus does not depend on the intractable $p_{\theta}(x)$. The Langevin dynamics (\ref{eq:langevin}) involves a gradient and a diffusion term. The first term is gradient descent ${z'_{k+1}=z'_k + s \frac{\partial}{\partial z} \log p_\theta(z_k|x)}$ on $\log p_\theta(z_k|x)$. If $z_k\sim p_\theta(z_k|x)$, then the distribution of $z'_k$ will be shifted towards basins of high log-posterior. We may recover $p_\theta(z_k|x)$ by smoothing with the second term $\sqrt{2s} \epsilon_k$, which amounts to white noise diffusion and induces randomness for sampling from $p_\theta(z_k|x)$. 

For small step size $s$, the marginal distribution of $z_k$ will converge to $p_\theta(z|x)$ as $k\rightarrow\infty$ regardless of the initial distribution of $z_0$~\cite{cover2012elements}. More specifically, let $q_k(z)$ be the marginal distribution of $z_k$ of the Langevin dynamics, then ${\rm KL}(q_k(z) \| p_\theta(z|x))$ decreases monotonically to 0, that is, by increasing $k$, we reduce ${\rm KL}(q_k(z) \| p_\theta(z|x))$. 

Finally, the MLE learning can be accomplished by gradient descent. Each learning iteration updates $\theta$ by 
\begin{align} 
\theta_{t+1} = \theta_t + \eta_t \frac{1}{n} \sum_{i=1}^{n} \E_{p_{\theta_t}(z_i|x_i)}\left[ \d \log p_{\theta}(x_i, z_i) \mid_{\theta = \theta_t} \right],  \label{eq:T0}
\end{align}
where $\eta_t$ is the step size or learning rate, and $\E_{p_{\theta_t}(z_i|x_i)}$  can be approximated by Monte Carlo sampling from $p_{\theta_t}(z_i|x_i)$.

\subsection{Learning with short run inference dynamics} 

It is computationally impractical to run long Markov chains from $p_\theta(z|x)$ as the gradient term in~(\ref{eq:langevin}) requires back-propagation through the model underlying $p_\theta(x|z)$. Earlier work~\cite{han2017abp} recruits persistent Markov chains~\citep{tieleman2008training} $\{(z_i, x_i), i=1,\ldots,n\}$ such that for each observed example $x_i$ a latent code $z_i$ is updated for a few steps in each learning iteration and the chains are maintained throughout the learning procedure. This method leads to inconsistent sampling procedures while training and evaluating the model, since persistent Markov chains for evaluation data are not available. Moreover, estimation of the log-likelihood has to resort to means such as annealed importance sampling~\cite{neal01ais}.

Instead, we adopt short run MCMC~\cite{nijkamp2019learning} in which we approximately sample from the posterior distribution of the latent variable. We thus propose the following short run inference dynamics, with a fixed small number of steps $K$~(e.g., $K=20$),
\begin{align} 
  & z_0 \sim p(z), \\
  & z_{k+1} = z_k + s \frac{\partial}{\partial z} \log p_\theta(z_k|x) + \sqrt{2s} \epsilon_k,  \label{eq:S}
  \end{align}
 where  $k = 1, ..., K$ and $p(z)$ is the prior distribution of $z$. Initializing $z_0\sim p(z)=\N(0, I)$, we perform $K$ steps of Langevin with step size $s$. 

Finally, the learning procedure updates $\theta$ by 
\begin{align} 
&\theta_{t+1} = \theta_t \\
&+ \eta_t \frac{1}{n} \sum_{i=1}^{n} \E_{q_{s, \theta_t}(z_i|x_i)}\left[ \d \log p_{\theta}(x_i, z_i) \mid_{\theta = \theta_t} \right],  \label{eq:T}
\end{align}
where $\eta_t$ is the learning rate, $\E_{q_{\theta_t}(z_i|x_i)}$ can be approximated by samples drawn from $q_{\theta_t}(z_i|x_i)$ using (\ref{eq:S}). Compared to MLE learning algorithm (\ref{eq:T0}), we replace $p_{\theta_t}(z|x)$ by $q_{s, \theta_t}(z|x)$. Moreover, we may update the step size $s$ of (\ref{eq:S}), which we will elaborate in the following.

\subsection{Theoretical understanding} 

Given $\theta_t$, the updating equation (\ref{eq:T}) is a one step gradient ascent on
\begin{align} 
Q_s(\theta) &=  \frac{1}{n} \sum_{i=1}^{n} \E_{q_{s, \theta_t}(z_i|x_i)}\left[ \log p_{\theta}(x_i, z_i) \right]. \label{eq:Qs}
\end{align}

Compared to the log-likelihood  in MLE learning, ${L(\theta) = \frac{1}{n} \sum_{i=1}^{n} \log p_\theta(x)}$, we have 
\begin{align} 
&Q_s(\theta) = L(\theta) + \frac{1}{n} \sum_{i=1}^{n}  \E_{q_{s, \theta_t}(z_i|x_i)}\left[ \log p_{\theta}(z_i|x_i) \right] \nonumber\\
&= L(\theta) - \frac{1}{n}\sum_{i=1}^{n}  {\rm KL}(q_{s, \theta_t}(z_i|x_i) \| p_{\theta}(z_i|x_i)) \nonumber\\
& +\frac{1}{n}\sum_{i=1}^{n}   \E_{q_{s, \theta_t}(z_i|x_i)}[\log q_{s, \theta_t}(z_i|x_i)]. \label{eq:Qs1}
\end{align}
Since the last term has nothing to do with $\theta$, gradient ascent on $Q_s(\theta)$ is equivalent to gradient ascent of 
\begin{align}
\tilde{Q}_s(\theta) &= L(\theta) - \frac{1}{n}\sum_{i=1}^{n}  {\rm KL}(q_{s, \theta_t}(z_i|x_i) \| p_{\theta}(z_i|x_i)), \label{eq:tQs} 
\end{align}
which is a perturbation or a variational lower bound of log-likelihood $L(\theta)$.

The fixed point of the learning algorithm (\ref{eq:T}) solves the following estimating equation: 
\begin{align}
\frac{1}{n} \sum_{i=1}^{n} \E_{q_{s, \theta}(z_i|x_i)}\left[ \d \log p_{\theta}(x_i, z_i)\right] = 0. \label{eq:E}
\end{align}
If we approximate $\E_{q_{s, \theta_t}(z_i|x_i)}$ by Monte Carlo samples from $q_{s, \theta_t}(z_i|x_i)$, then the learning algorithm becomes Robbins-Monro algorithm for stochastic approximation \cite{robbins1951stochastic}, whose convergence to the fixed point follows from regular conditions of Robbins-Monro. The estimating equation (\ref{eq:T}) is a perturbation of the maximum likelihood estimating equation $\frac{1}{n} \sum_{i=1}^{n} \E_{p_{\theta}(z_i|x_i)}\left[ \d \log p_{\theta}(x_i, z_i)\right] = 0$. 

\subsection{Optimizing step size} 

We can optimize the step size $s$ by maximizing $\tilde{Q}_s(\theta)$ defined in equation (\ref{eq:tQs}), which is equivalent to minimizing the KL divergence between the short-run-dynamics-induced posterior and the true posterior since the first term $L(\theta)$ does not involve $s$. $\tilde{Q}_s(\theta)$ involves the entropy of $q_{s, \theta_t}(z_i|x_i)$. We provide the details of its computation in the supplementary materials. The step size optimization can be done by grid search or stochastic gradient descent. In this work, we optimize the step size $s$ with grid search guided by maximizing $\tilde{Q}_s(\theta)$.

\subsection{Algorithm} 

The learning procedure is summarized in Algorithm~\ref{algo:short}. Note that we only optimize $s$ every $T_s$ iterations, so that computational cost is negligible.

\begin{algorithm}[h!]
	\SetKwInOut{Input}{input} \SetKwInOut{Output}{output}
	\DontPrintSemicolon
	\Input{Learning iterations~$T$, step size interval $T_s$, learning rate~$\eta$, initial weights~$\theta_0$, observed examples~$\{x_i \}_{i=1}^n$, batch size~$m$, number of steps $K$, initial step size $s$.}
	\Output{Weights $\theta_{T+1}$.}
	\For{$t = 0:T$}{			
		\smallskip
		1. Draw observed examples $\{ x_i \}_{i=1}^m$. \;
		2. Draw latent vectors $\{ z_{i,0} \sim p(z) \}_{i=1}^m$.\;
		3. Infer $\{ z_{i,K} \}_{i=1}^m$ by $K$-steps of dynamics (\ref{eq:S}) with step size $s$.\;
		4. Update $\theta$ according to (\ref{eq:T}).\;
		5. Every $T_s$ iterations, update $s$.\;
	}
	\caption{Learning with SRI.}
	\label{algo:short}
\end{algorithm}

\subsection{Log-likelihood computation} \label{likelihood}

Unlike traditional MCMC, short run inference enables the computation of the marginal log-likelihood $\log p(x)$\footnote{Note that its Monte Carlo estimator is biased but the bias is diminishing with a large sample size.},
\begin{align}
\label{eq:logp}
\log p_\theta(x)
&= \log\int p_\theta(x,z) dz \nonumber\\
&= \log\int\frac{p_\theta(x|z)p(z)}{q_k(z)}q_k(z)dz \nonumber\\
&= \log E_{q_k(z)}\left[\frac{p_\theta(x|z)p(z)}{q_k(z)}\right].
\end{align}
Then,
\begin{align}
\label{eq:logp2}
&E_{p_{\rm data}}\left[\log\frac{1}{M}\sum_{i=1}^M\frac{p_\theta(x|z_i)p(z_i)}{q_k(z_i|x)}\right]\nonumber\\
&=E_{p_{\rm data}}\left[\log\sum_{i=1}^{M}\exp\left[\log p_\theta(x|z_i)\right.\right.\nonumber\\
&\left.+\log p(z_i)-\log q_k(z_i|x)\right] - \log M\Biggr].
\end{align}
While most terms in (\ref{eq:logp2}) are readily available, $\log q_k(z_i|x)$ requires special treatment. We may rewrite the dynamics~(\ref{eq:S}) in the form of 
\begin{align}
\label{eq:sr}
z_0 \sim p(z), \quad z_k = R_k(z_0)
\end{align}
where $R_k$ is defined by a $k$-step Langevin dynamics. Let the distribution of $z_k$ be denoted $q_k(z)$. Then, by change of variable theorem,
\begin{align} 
&z_k\sim q_k(z),\\
&q_k(z) = p(R_k^{-1}(z))|{\rm det}(d R_k^{-1}(z)/d z)|. 
\end{align}
Instead of inverting $R_k$, we draw $z_0\sim p(z)$ and compute the log determinant of the Jacobian $d R_k(z_0)/d z_0$. See more details in the supplementary. 

\section{Related Work}

{\em Variational inference.} VAE \cite{kingma2013auto,bowman-etal-2016-generating} is a prominent method for learning generative models. Due to the autoregressive nature of the decoder, a naive application of VAE to text data results in posterior collapse. Following work makes extensive efforts to alleviate this issue \citep{fu2019cyclical, yang2017dilated, he2018lagging, li2019surprisingly, kim2018semi, pelsmaeker-aziz-2020-effective, dieng2019reweighted}. Among them SA-VAE developed by \citet{kim2018semi} is mostly related to our work. They propose SA-VAE where initial variational parameters obtained from the inference model are further refined by running a small number of gradient updates (e.g., 20) guided by the ELBO. In our work, instead of relying on a parameteric varational distribution, we run a few gradient updates on the log-posterior of the latent variable with initialization from the prior distribution to draw samples directly. Thus, there is no need to design and tune an extra inference model, which is highly non-trivial considering that posterior collapse occurs easily in VAE training.

{\em Alternating back-propagation.} \citet{han2017abp} propose to learn generative models for images by maximum likelihood, where the learning algorithm iterates over two steps: (i) inferring the latent variable by sampling from its posterior distribution with Langevin dynamics; (ii) updating the model parameters based on the inferred latent codes. In the training stage, in step (i), the Langevin dynamics is initialized from the latent codes inferred in the last epoch, which is called persistent chain in the literature~\citep{tieleman2008training}. In contrast, the short run dynamics always initializes the gradient descent updates from the prior noise distribution. Data-independent initialization renders the dynamics in training and testing consistent.   

{\em Short run MCMC.} \citet{nijkamp2019learning} introduces short run MCMC as a learned sampling dynamics guided by an energy-based model. It shares the same theoretical underpinning as early work of using stochastic gradient Langevin dynamics to learn mixture of Gaussians and logistic regression for large-scale data \cite{welling2011bayesian}. Our short run inference method for learning latent variable models for text is inspired by these works. Simultaneous to this paper, \citet{nijkamp2020learning} applied short run inference to multi-layer top-down generative models for images.  

\begin{table}[t]
	\begin{center}
		\begin{tabular}{ccccc}
			\specialrule{.1em}{.05em}{.05em}
			& PPL & Recon & AU & KL \\
			\hline
			\multicolumn{5}{c}{PTB}\\
			LSTM-LM & 100.47 & - & - & - \\
			Anneal-VAE & 101.40 & 101.28 & 0 & 0.00 \\
			Cyclical-VAE & 108.97 & 101.85 & 5 & 1.37 \\
			Lagging-VAE & 99.83 & 100.26 & 4 & 0.93 \\
			SA-VAE & 100.39 & 100.97 & 5 & 1.86 \\
			FBP-VAE & 99.62 & 98.52 & 3 & 2.95 \\
			FB-VAE & 96.35 & 94.52 & 32 & 8.15 \\
			Ours & $\mathbf{94.26}$ & $\mathbf{91.14}$ & $\mathbf{32}$ & $\mathbf{10.13}$ \\
			\hline
			\multicolumn{5}{c}{SNLI}\\
			LSTM-LM & 21.44 & - & - & - \\
			Anneal-VAE & 21.50 & 31.66 & 2 & 1.42 \\
			Cyclical-VAE & 21.62 & 30.89 & 4 & 2.36 \\
			Lagging-VAE & $\mathbf{21.16}$ & 31.53 & 5 & 1.42 \\
			SA-VAE & 21.49 & 30.12 & 5 & 2.34 \\
			FBP-VAE & 21.46 & 31.04 & 3 & 2.12 \\
			FB-VAE & 22.00 & 23.36 & 32 & 8.48 \\
			Ours & 21.21 & $\mathbf{22.24}$ & $\mathbf{32}$ & $\mathbf{10.02}$ \\			
			\hline
			\multicolumn{5}{c}{Yahoo}\\
			LSTM-LM & 60.75 & - & - & - \\
			Anneal-VAE & 61.52 & 329.10 & 0 & 0.00 \\
			Cyclical-VAE & 66.93 & 333.80 & 0 & 2.83 \\
			Lagging-VAE & 59.77 & 322.70 & 15 & 5.70 \\
			SA-VAE & 63.92 & 327.27 & 17 & 7.23 \\
			FBP-VAE & 62.88 & 328.13 & 2 & 3.06 \\
			FB-VAE & 59.51 & 315.31 & 32 & 15.02 \\
			Ours & $\mathbf{57.05}$ & $\mathbf{311.23}$ & $\mathbf{32}$ & $\mathbf{16.19}$ \\						 
			\specialrule{.1em}{.05em}{.05em}
		\end{tabular}
	\end{center}
	\caption{Language modeling results on PTB, SNLI, and Yahoo test set.}
	\label{tab:language_model}
\end{table}

\section{Experiments}
We apply our method to train latent variable models on text datasets. The dimension of the latent variable is $32$ in all experiments. The generator is implemented with a one-layer uni-directional LSTM \citep{hochreiter1997lstm}. The number of hidden units and word embedding size of the LSTM vary among datasets to closely follow the experimental setup in recent work \citep{fu2019cyclical, li2019surprisingly}. The number steps of the short run dynamics is $20$ for all experiments \footnote{$K=10$ steps led to posterior collapse. We observed a slight improvement in model performance if $K$ was increased from 20 to 40 and no improvement from 40 to 60.}. The sample from the short run dynamics is used to predict the initial hidden state of the LSTM. It is also concatenated with the word embeddings and then fed to the LSTM as input at each time step.

The short run inference is more computationally costly than the vanilla VAE and has comparable training cost as some improved versions of VAE. The number of inner steps of SRI (20 steps) is about the same as that of SA-VAE and Lagging-VAE. In training, SRI has faster convergence than SA-VAE and comparable convergence as Lagging-VAE in our experiments. In inference, our sampling-based approach is slower than amortized inference. Our method trades a feasible computational cost for accurate inference whose empirical performance is presented in the following experiments \footnote{Our implementation is available \url{https://github.com/bpucla/sri_text}}.

\begin{table}
	\begin{center}
		\begin{tabular}{l}
			\specialrule{.1em}{.05em}{.05em}
			\multicolumn{1}{c}{\textbf{FB-VAE}}\\
			a man with a cane is walking down the street . \\
			a man with a cane is walking down the street . \\
			a man in a blue shirt is eating food . \\
			people are eating food . \\
			people walk in a city . \\
			people are outside in a city . \\
			\hline
			\multicolumn{1}{c}{\textbf{Ours}}\\
			there is a boy skating down a small street .\\
			there is a child walking in the snow .\\
			the man is riding a horse through the snow .\\
			the man is riding a boat .\\
			the biker is looking at the lake .\\
			the person is looking at a country .\\
			\specialrule{.1em}{.05em}{.05em} 
		\end{tabular}
	\end{center}
	\caption{Comparison on interpolation. Sentence samples greedily decoded from linear interpolation between samples from the Gaussian prior with FB-VAE (Top) and SRI-trained generative model (Bottom).}
	\label{tab:interpolation}
\end{table}

\begin{table*}
	\begin{center}
		\begin{tabular}{l}
			\specialrule{.1em}{.05em}{.05em}
			there is a crowd of people in the city . \\
			a man rubbing a dirty face . \\
			a couple was waiting to cross the street in a grocery store . \\
			the little girl is drinking water . \\
			a group of boys are playing in the fountain \\
			\hline
			five asian teenagers are peforming a dance routine for a volunteer organization .\\
			a white-haired man is in front of a building playing music .\\
			construction workers sit at a african courtyard .\\
			a jewish man wearing white garb , playing a guitar , with a lazy look on him .\\
			a young man in a brown checkered shirt sings down on the floor while playing with the on a hot day .\\
			\specialrule{.1em}{.05em}{.05em} 
		\end{tabular}
	\end{center}
	\caption{Comparison on the generated sentences. Sentence samples generated from the Gaussian prior by FB-VAE (Top) and SRI-trained generative model (Bottom).}
	\label{tab:generation}
\end{table*}

\begin{table}
	\begin{center}
		\begin{tabular}{ccccc}
			\specialrule{.1em}{.05em}{.05em}
			$k$ & $1$ & $2$ & $3$ & $4$\\
			\hline
			AE & \textbf{26.05} & $40.46$ & $52.77$ & $63.07$ \\
			Anneal-VAE & $32.20$ & $32.65$ & $33.12$ & $33.39$ \\
			Cyclical-VAE & $31.83$ & $32.87$ & $33.73$ & $34.38$ \\
			Lagging-VAE & $31.78$ & $31.99$ & $32.21$ & $32.32$ \\
			SA-VAE & $31.63$ & $31.82$ & $32.15$ & $32.46$ \\
			FBP-VAE & $29.93$ & $32.59$ & $34.90$ & $36.77$ \\
			FB-VAE & $27.92$ & $29.12$ & \textbf{30.03} & $30.85$ \\
			Ours & 27.12 & \textbf{28.66} & 30.21 & \textbf{30.46} \\
			\specialrule{.1em}{.05em}{.05em} 
		\end{tabular}
	\end{center}
	\caption{Noisy reconstruction loss on SNLI. $k$ is the number of word swaps performed on the original sentences.}
	\label{tab:noisyrecon}
\end{table}

\begin{table}
	\begin{center}
		\begin{tabular}{ccccc}
			\specialrule{.1em}{.05em}{.05em}
			Number of Labels & $0$ & $100$ & $1k$ & $10k$\\
			\hline
			AE & $53.1$ & $78.8$ & $83.7$ & $84.1$ \\
			Anneal-VAE & $56.3$ & $59.2$ & $62.3$ & $65.8$ \\
			Cyclical-VAE & $59.9$ & $78.6$ & $82.7$ & $83.2$ \\
			Lagging-VAE & $63.6$ & $65.8$ & $74.2$ & $80.5$ \\
			SA-VAE & $62.6$ & $69.3$ & $78.8$ & $81.4$ \\
			FBP-VAE  & $60.9$ & $74.8$ & $76.9$ & $81.1$ \\
			FB-VAE  & $67.5$ & $84.9$ & $89.5$ & $90.6$ \\
			Ours & \textbf{73.3} & \textbf{85.8} & \textbf{89.6} & \textbf{90.8} \\
			\specialrule{.1em}{.05em}{.05em} 
		\end{tabular}
	\end{center}
	\caption{Accuracy on Yelp of unsupervised and semi-supervised classification as a function of the number of labeled example during training.}
	\label{tab:classification}
\end{table}

\subsection{Language Modeling}
We evaluate our method on language modeling with the Penn Tree Bank (PTB) \citep{marcus1993building}, Yahoo \citep{yang2017dilated}, and a downsampled version of the Stanford Natural Language Inference (SNLI) corpus \citep{bowman-etal-2015-large} as preprocessed in \citep{li2019surprisingly}. Ideally, a language model with latent variable would be expected to make use of the latent space and accurately model the data distribution. To measure the utilization of the latent space, three quantitatively metrics are often considered in prior work \citep{bowman-etal-2016-generating,li2019surprisingly,fu2019cyclical,kim2018semi}: reconstruction error (Recon), number of active units (AU), the magnitude of KL. Reconstruction error is the negative log-likelihood of the observed data evaluated under the posterior, $\E_{q(z|x)}\left[- \log p_{\theta}(z|x) \right]$. A latent dimension is considered active if its distribution changes depending on the observations. Following \citet{burda2015importance}, a latent dimension is defined to be active if $\text{Cov}_{x} (\E_{z \sim q(z | x)}[z]) > 10^{-2}$. Perplexity (PPL) based on the marginal log-likelihood of $x$ is adopted to measure how accurately the model captures the data. The marginal log-likelihood is estimated with importance sampling with $z$ samples from trained short run dynamics as importance samples. 

Besides the standard LM and the vanilla VAE with KL weight annealing, VAEs with recent state-of-the-art training techniques, Cyclical-VAE \citep{fu2019cyclical}, Lagging-VAE \cite{he2018lagging}, SA-VAE \cite{kim2018semi}, FBP-VAE and FB-VAE \cite{li2019surprisingly} are also included for comparison. The results are displayed in Table \ref{tab:language_model}. In terms of PPL, our method outperforms all the baselines on the PTB and Yahoo datasets, while does slightly worse than Lagging-VAE and performs better than other baselines on the SNLI. This indicates the model trained with our method is able to accurately model the data distribution. On the other hand, our method yields the lowest reconstruction error and the highest KL with all latent dimension active on all three datasets, exposing the active use of the latent space. Taken together, these results suggest that the model trained with short run dynamics are balanced on modeling the data and utilizing the latent space.

\begin{figure}[h]
\begin{center}
	\begin{subfigure}{.5\textwidth}
		\centering
		\includegraphics[width=0.6\linewidth]{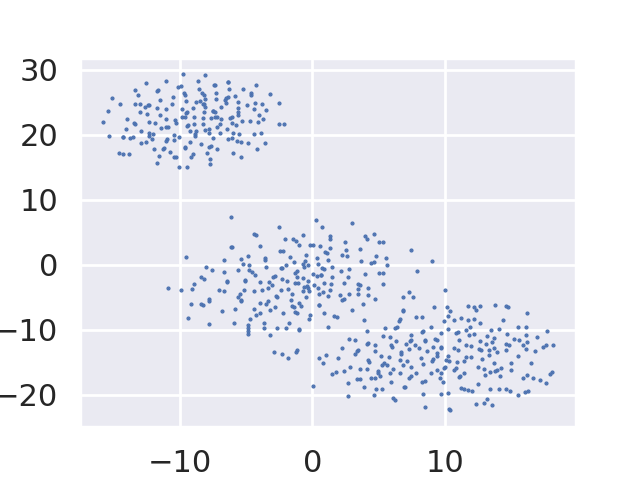}
		\vspace{0.05cm}
	\end{subfigure}
	\begin{subfigure}{.5\textwidth}
		\centering
		\includegraphics[width=0.8\linewidth]{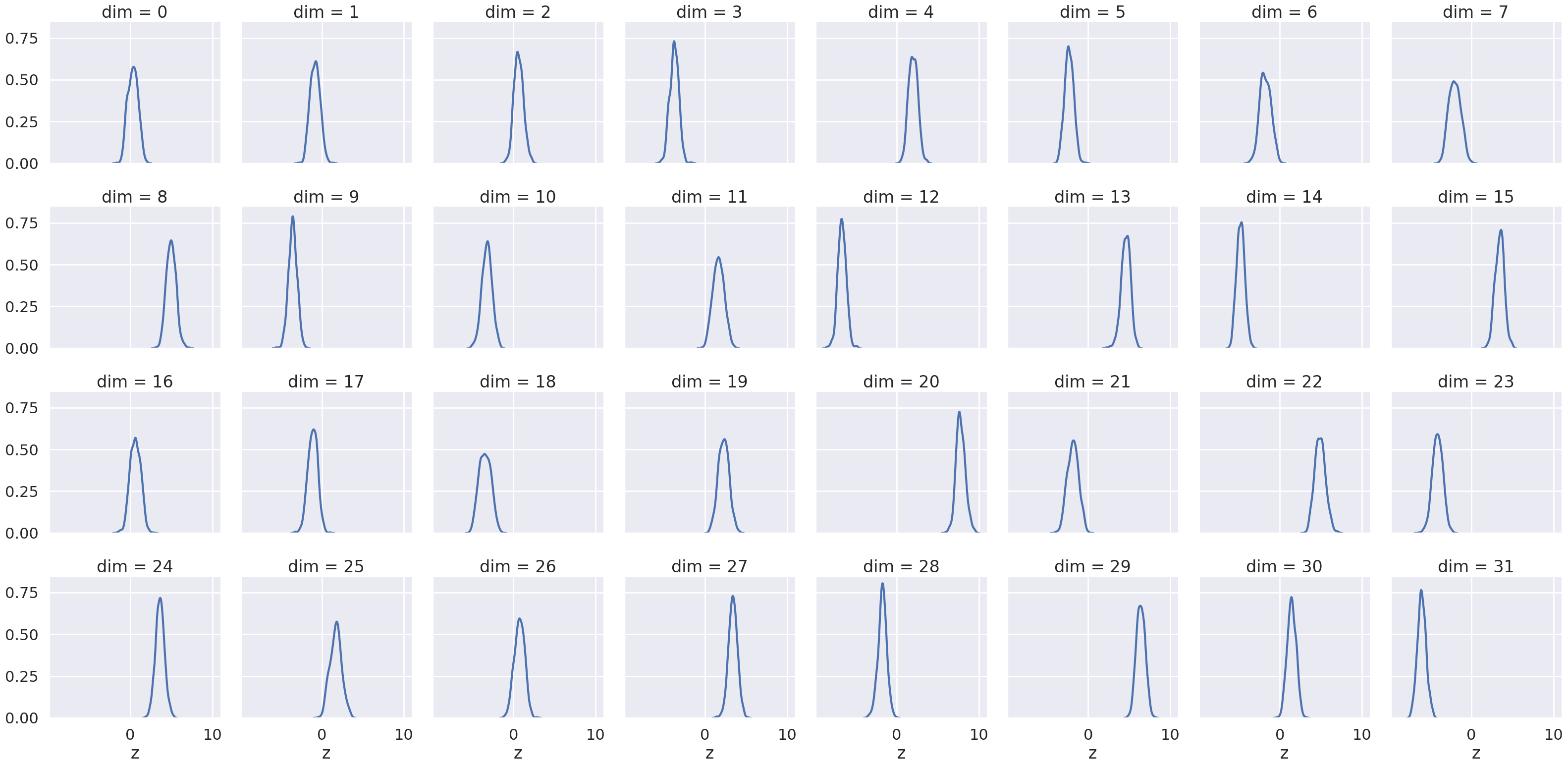}
		\vspace{0.05cm}
	\end{subfigure}
	\caption{\small t-SNE plot (upper) and marginal density plot (bottom) of the SRI-induced aggregate posterior on the PTB dataset.}
	\label{fig:tsne}
	\end{center}
\end{figure}

Figure~\ref{fig:tsne} displays a t-SNE plot of the SRI-induced aggregate posterior $E_{p_{\rm data}} q (z|x)$ and its marginal density of each dimension. The t-SNE plot demonstrates the SRI-induced aggregate posterior is multi-modal and the marginal densities are uni-modal but clearly deviates from the zero-centered standard Gaussian prior. These visualizations demonstrate that the aggregate posterior in our model is clearly different the isotropic Gaussian prior\footnote{Ideally $E_{p_{\rm data}} q (z|x) = p(z)$ since $\int_x p_{\rm data}(x) q(z|x) = \int_x p(z) p_\theta(x|z)$. However the generative model might not be able to induce such a model posterior. The mismatch might indicate some form of under-regularization, similar to other approaches for mitigating posterior collapse such as FB-VAE.} and thus our model does not show a posterior collapse issue, consistent with our analysis above. 

\subsection{Latent Space Analysis}
The quality of the latent space with SNLI is examined through interpolation, generation, and noisy reconstruction. 

\subsubsection{Interpolation}
Interpolation allows us to appraise the smoothness of the latent space. In particular, two samples $z_1$ and $z_2$ are drawn from the prior. We linearly interpolate between them and then decode the interpolated points. FB-VAE \citep{li2019surprisingly} is considered as the SOTA text VAE that mitigates posterior collapse. Due to space limit, we only include this method for comparison in interpolation and generation experiments. Table \ref{tab:interpolation} shows the decoded samples. Although the interpolated sentences by FB-VAE appears smooth, the first two sentences are repetitive. In comparison, the decoded sentences from our model transition more smoothly. While the interpolated sentences from our model are diverse, their syntactic properties and topic information remain consistent in neighborhoods along the path, exposing a smooth latent space.  

\subsubsection{Generation}
We sample from the prior distribution and decode the sentences in a greedy manner. Table~\ref{tab:generation} displays the samples from our model and FB-VAE. It appears that samples from both models are grammatically correct and semantically meaningful in general. FB-VAE samples nevertheless show more local grammar errors. More generated samples are given in the supplementary. 

\subsubsection{Noisy reconstruction}
\citet{zhao2018adversarially} reasons that a latent variable model's capacity on reconstructing from noisy data reveals the smoothness of the latent space. We impose discrete noise to the data by swapping tokens in a sentence for $k$ times, where $k = 1, 2, 3, 4$ in this experiment. The reconstruction error (negative log-likelihood) under each condition is reported in Table \ref{tab:noisyrecon}. Notice that even the AE yields the lowest reconstruction when the noise is low ($k = 1$), but its performance deteriorates quickly as the noise level increases, implying that the latent space of AE is not smooth. In contrast, other models with regularization on the latent space do not exhibit drastic decline in reconstruction performance with increasing noise level. Furthermore, the model trained with our method demonstrates reconstruction either outperforming other methods or comparable to the best, revealing that the model trained with SRI has a smooth latent space. 

\subsection{Classification}
The latent space of a well-learned latent variable model should capture highly informative features such that data points cluster into meaningful groups in the latent space. We hence further probe the latent space structure by investigating the clustering and classification performance of the SRI-inferred latent codes. Following prior work \citep{fu2019cyclical, li2019surprisingly}, we utilize the Yelp sentiment dataset as preprocessed in \citet{shen2017style}. We train a Gaussian mixture for clustering (zero labels) and a SVM with 100, 1000, or 10,000 number of labels. The results are displayed in Table \ref{tab:classification}. Our method consistently improves over VAE approaches and AE. The improvement is especially clear in the zero-shot setting and small data regime (0 and 100 labels), revealing a well-structured latent space learned by SRI.

\section{Conclusion}
This work proposes to use short run inference dynamics to infer latent variables in text generative models. SRI dynamics is always initialized from the prior distribution of the latent variable and then performs a small number (e.g., 20) of Langevin dynamics updates guided by the posterior distribution. This simple and automatic inference method induces a good approximate posterior and provides good latent code.

The model trained with SRI accurately models the text data compared to strong language model and generative model baselines and shows no sign of posterior collapse, which is non-trivial to avoid and several remedies have been proposed for in prior art. Moreover, the learned space is smooth and captures rich representations of the sentences.

\section*{Acknowledgement} 

We thank the reviewers for their insightful comments and suggestions. The work is supported by NSF DMS-2015577.

\bibliography{anthology,eacl2021}
\bibliographystyle{acl_natbib}

\end{document}